\documentclass[a4paper, 10pt, conference]{ieeeconf}      

\IEEEoverridecommandlockouts                              

\usepackage{amsmath} 
\usepackage{amssymb}  
\usepackage{graphicx}
\usepackage{tabu}
\usepackage{diagbox}
\usepackage{xcolor}
\usepackage{soul}
\usepackage{caption}
\usepackage{booktabs}

\title{\LARGE \bf
Good Time to Ask: A Learning Framework\\for Asking for Help in Embodied Visual Navigation
}

\author{Jenny Zhang$^{1}$, Samson Yu$^{2}$, Jiafei Duan$^{3}$ and Cheston Tan$^{4}$%
\thanks{$^{1}$Author is with University of British Columbia, Canada, {\tt\small jennyzzt@cs.ubc.ca}}%
\thanks{$^{2}$Author is with National University of Singapore, Singapore}%
\thanks{$^{3}$Author is with University of Washington, United States of America}%
\thanks{$^{4}$Author is with Centre for Frontier AI Research, A*STAR, Singapore}%
}

\begin{document}

\maketitle
\thispagestyle{empty}
\pagestyle{empty}

\begin{abstract}

In reality, it is often more efficient to ask for help than to search the entire space to find an object with an unknown location. We present a learning framework that enables an agent to actively ask for help in such embodied visual navigation tasks, where the feedback informs the agent of where the goal is in its view. To emulate the real-world scenario that a teacher may not always be present, we propose a training curriculum where feedback is not always available. We formulate an uncertainty measure of where the goal is and use empirical results to show that through this approach, the agent learns to ask for help effectively while remaining robust when feedback is not available.

\end{abstract}

\section{Introduction}

\begin{figure}[ht]
    \centering
    \includegraphics[width=1.0\linewidth]{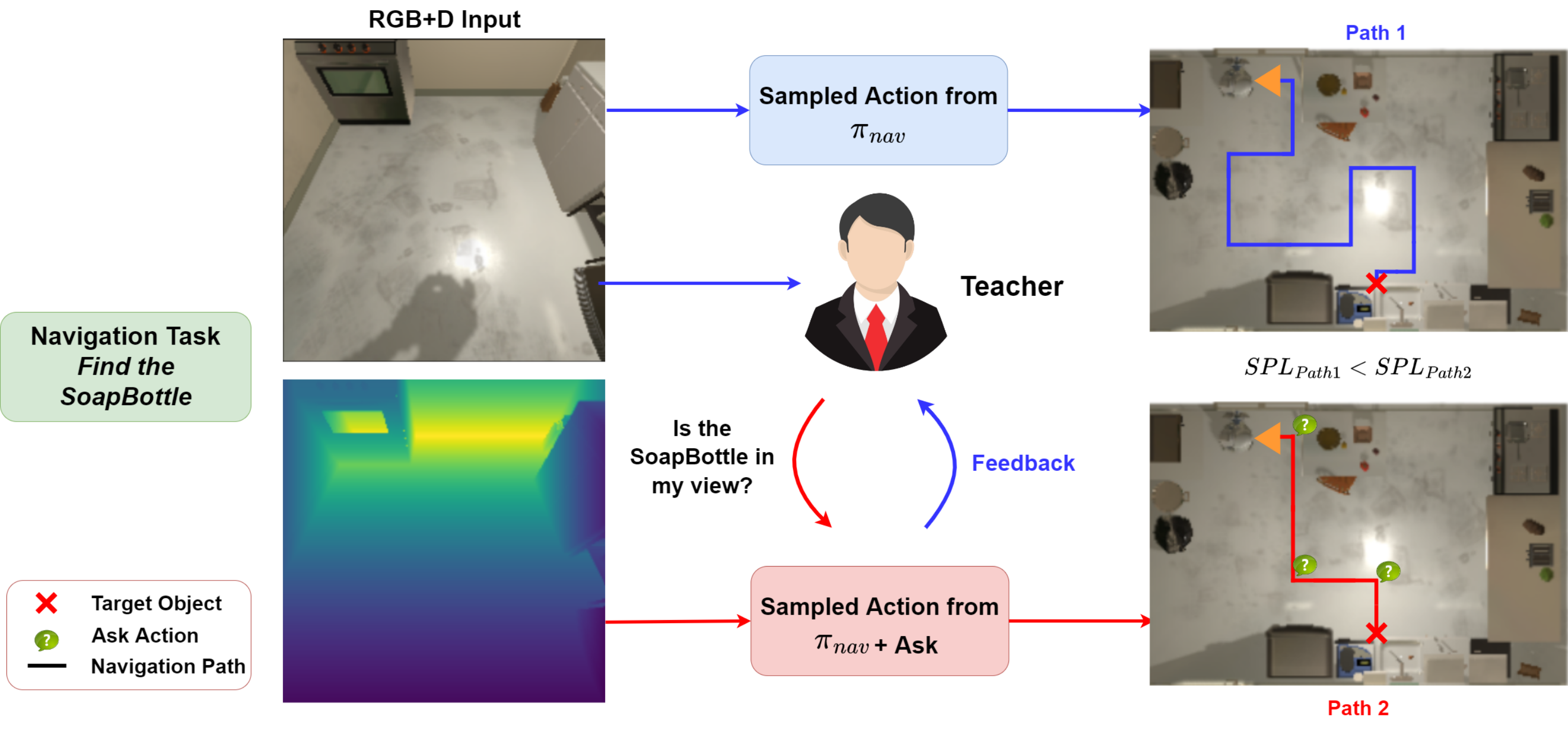}
    \caption{Our learning framework where the agent has an additional capability to ask for feedback during ObjectNav. The overall pipeline of our learning framework and an illustrated comparison in the ObjectNav performance between agents with and without the \emph{ask} action.}
\label{fig:0}
\vspace{-5pt}
\end{figure}

Consider the following scenario: you instruct a newly deployed robot assistant to retrieve a tool for you, but the robot does not know its location. Hence, it searches your entire home for it. It would be far more efficient if it could ask someone familiar with the task and environment for help, as shown in Fig. \ref{fig:0}.
We naturally want to seek assistance when a task is difficult, especially so when we lack sufficient information about the goal or the environment. Similarly, robots should be endowed with the ability to actively query and gain relevant knowledge. However, agents in many robotics works \cite{anderson2018vision,chen2019touchdown,gadre2022clip} do not have access to external assistance when tackling their navigation tasks.

We propose \textbf{Good Time to Ask (GTA)}, a learning framework to train and evaluate an embodied agent with the additional capability of asking for feedback. The task is set to be object-goal navigation (ObjectNav) in AI2-THOR \cite{kolve2017ai2}, an embodied AI \cite{weihs2020allenact} simulator with photorealistic indoor environment. An interactive reinforcement learning approach is used to train the agent to learn to seek help in the form: ``Is the target object in view?". To evaluate the quality of a query by the agent in the ObjectNav task, we present two important considerations:
\begin{itemize}
  \item \emph{Timing} - When is the best time to ask for assistance?
  \item \emph{Robustness} - How to prevent the agent from being overly reliant on feedback?
\end{itemize}

The agent must learn to ask for help in a timely way that maximizes the feedback's informativity while minimizing inconvenience (i.e. annotation cost) to the teacher. 
In addition, it is important to account for the possibility that a teacher is not present to improve the generalizability of our setup to real-world settings. Hence, we need to ensure that the agent is not overly reliant on feedback and can still perform well when external help is unavailable.
We develop a novel metric to assess the quality of a query and address the considerations above. This is done by calculating the lower bound estimate of the agent's uncertainty with regards to the task objective and the type of action taken by the agent at each time step.

The main contributions of our proposed learning framework are threefold: 1) We demonstrate the positive effect of teacher-supplied assistance, in the form of image semantic segmentation feedback, on the performance of learning agents. 2) We propose a semi-present teacher training curriculum to train agents that can adapt their behavior to the teacher's level of availability. 3) We propose a new evaluation metric to assess the quality of the agent's queries for help. 

\section{Related Works}

\textbf{Object-Goal Navigation} \cite{zhu2017target, batra2020objectnav, zhu2021deep, gadre2022clip} has been intensively studied as a fundamental task in the field of embodied AI \cite{weihs2020allenact, eaisurvey, alfred} in recent years. In its simplest form, ObjectNav is the task of navigating to an object in photorealistic 3D environments \cite{kolve2017ai2, savva2019habitat, li2021igibson}.
Since ObjectNav is a sequential decision-making problem, existing learning-based approaches rely on reinforcement learning (RL) and can be modeled as a Markov Decision Process. Learning-based approaches for ObjectNav differ in their memory architectures, e.g. gated recurrent units \cite{cho2014properties}, semantic maps \cite{chaplot2020object}, and amount of environment prior exploration \cite{ramakrishnan2021exploration}.
In this work, we focus on making key modifications to the RL framework by closing the loop with human feedback.

\textbf{Interactive Reinforcement Learning} \cite{poole2021towards, arzate2020survey} accounts for human guidance and feedback \cite{zhang2019leveraging, 10.3389/frobt.2021.584075, maclin1996creating}.
Interactive RL adopts a human-in-the-loop \cite{wu2022survey} approach to integrate contextual human knowledge that improves or personalizes the behavior of AI agents. This helps to reduce the notorious issue of sample inefficiency in RL \cite{6005223}. The incorporation of human advice into RL is commonly done through reward shaping \cite{zhang2019leveraging,arzate2020survey}.
An alternative approach uses human advice as input observations for the RL agent's policy, e.g. gestures in Ges-THOR \cite{wu2021communicative} and state descriptions \cite{nguyen2021learning}. Lastly, human advice can be expressed in diverse ways e.g. language \cite{dialfred, teach}, gesture \cite{wu2021communicative}, and image \cite{poole2021towards, nguyen2021learning}. We focus on human feedback in the form of image semantic segmentation, which is used as part of the agent's observation. Compared to previous work \cite{wu2021communicative, chi2020just, thomason2019}, this type of human feedback is less costly as it does not require the teacher to be always physically next to the agent and requires less environment knowledge.

\textbf{Active Learning} attempts to maximize an agent’s task performance while minimizing the amount of samples annotated \cite{ren2021survey, 7bb36d742bf34b9d854aa0778ec47903}. Similarly, we aim to minimize the AI agent's requests for human feedback while maximizing the agent's navigation performance. In general, there are rule-based and learning-based approaches for active learning in Interactive RL. In rule-based approaches, pre-defined heuristics determine when to ask for external help. For example, in the model-confusion method, the agent asks for guidance on the best next action in vision-and-language navigation \cite{gu2022vision} when the difference in the policy's top two action probabilities exceeds a threshold \cite{chi2020just}. In learning-based approaches, the agent learns when to ask for help \cite{chi2020just, nguyen2019help, zhang2023robustness}. Previous work has found learning-based approaches to be more robust than rule-based ones \cite{chi2020just, tan2023perceptive}. Here, we use a learning-based approach and expand the action space in our agent's policy to accommodate the \emph{ask} action in the ObjectNav task.

\section{Methods}

\subsection{Problem Formulation}
In this work, we tackle the ObjectNav task \cite{batra2020objectnav}. To complete the task, the agent must navigate to the target object instance with stopping distance $\leq$ 1.0m, and then issue a termination (i.e. \textit{stop}). The object must be within the agent's field of view in order to succeed. An episode is terminated if the agent issues a termination action, regardless of whether it has succeeded, or if the maximum allowed time step is reached, which is 500 in our setup. Two key differences with \cite{batra2020objectnav} is that the target object is chosen randomly and the placement of objects is randomized every episode. There is only one instance of the target object type in every episode, and is guaranteed to be reachable. There are 7 object categories available for all scenes. However, only 5 objects (apple, bowl, potato, soap bottle, and dish sponge) are used in training while we reserve 2 objects (cup and bread) for unseen object test scenarios.

We choose AI2-THOR as our learning environment to train and evaluate our embodied agent because it has diverse scenes and interactive features. AI2-THOR has been widely used for different visual navigation tasks \cite{yang2018visual, zhu2017target, alfred}. It also allows the possibility of deploying our learning framework into a real-world robot via RoboTHOR \cite{RoboTHOR}. We use 10 kitchen scenes for training and 5 kitchen scenes for testing. Each scene has its own unique appearance and arrangement. The scene used for training is randomized every episode. Each training experiment is ran for 10 million steps and evaluated for 100 episodes on each scene and object.

The learning agent is represented by a capsule-shaped robot character in AI2-THOR. The agent has six available navigation actions: [\textit{rotate left}, \textit{rotate right}, \textit{move forward}, \textit{look up}, \textit{look down}, and \textit{stop}]. In setups where the feedback mechanism is enabled, the agent has an additional action: [\textit{ask}]. Each rotation action results in a 90\textdegree\ rotation; each look up or down action results in an 30\textdegree\ increment or decrement in the agent's view angle; each forward action results in a 0.25m forward displacement. The agent is initialized at a fixed location in each scene.

\subsection{Model Architecture}

\begin{figure}[ht]
    \centering
    \includegraphics[width=0.9\linewidth]{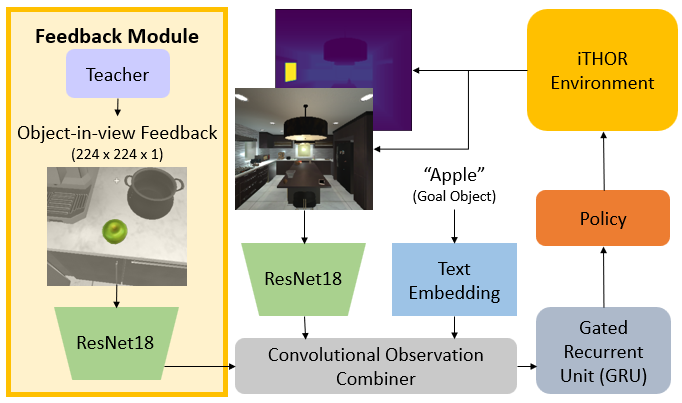}
    \caption{Overall architecture used to learn ObjectNav with the example of a target object ``Apple''. The highlighted feedback module on the left is not present in the baseline model.}
\label{fig:architecture}
\vspace{-5pt}
\end{figure}

Fig. \ref{fig:architecture} shows an overview of the model architecture used. Like most embodied AI works, we equip our agent with a RGB-D sensor. These visual observations are encoded with pretrained ResNet-18 models \cite{resnet}. The target object is encoded into a text embedding. In addition to these classic observations, we provide an \textbf{object-in-view observation} that is provided upon each \textit{ask} action, containing the ground truth semantic segmentation observation of the target object's location in the agent's view. Pixels which correspond to the target object's location have a value of 1, and 0 otherwise. The combined observations are passed into a gated recurrent unit (GRU) \cite{gru} before the RL model.

We formulate our ObjectNav learning framework using deep RL, specifically an on-policy actor-critic reinforcement learning algorithm - PPO \cite{ppo}. The actor-critic model has a shared backbone consisting of a 2-layer network, each layer having 256 nodes and ReLU activation. Linear prediction heads are used to obtain the value estimate and action distribution. 
We implement PPO with a time horizon of 128 steps, batch size of 128, discount factor of 0.99, 4 epochs for each iteration of gradient descent, and buffer size of 2048 for each policy update. We use Adam \cite{adamoptimizer} as the optimizer with a learning rate of $3\mathrm{e}{-4}$. The agent receives a positive reward of +10 if it completes the navigation successfully. To encourage the agent to reach the target object in the minimum number of steps, the agent receives a small penalty of -0.01 for each time step. AllenAct \cite{weihs2020allenact, wijmans2019dd} is used as the codebase for our framework.

\subsection{Training Curriculum with Semi-Present Teacher}
\label{sec:semi-present-teacher}
While the agent should learn to make use of external feedback when available, the teacher may not always be present to provide assistance in a real-world setting. When help is unavailable, the desired behavior for the agent in this task would be to navigate the scene autonomously to find the target object, even if it takes more time.

Hence, we introduce a semi-present teacher training curriculum to improve the agent's robustness in both settings. We compare between two training curricula: 25\% and 75\% semi-present teacher. A $\eta$\% semi-present teacher is present in $\eta$\% of training episodes. The agent has an additional observation of whether the teacher is present. Feedback is only available to the agent when the teacher is present.

\subsection{Quantifying Uncertainty}
\label{sec:quantifying-uncert}
We propose a metric to quantify the agent's uncertainty of where the goal is, which is the target object in this case. This metric is also used to quantitatively assess the impact of the \textit{ask} actions. We use $\mathbb{N}$ and $\mathbb{R}$ to denote the set of natural and real numbers respectively.
In a scene~$s$, let $P_s \subseteq \langle \mathbb{N} \times \mathbb{N} \times \mathbb{N} \rangle$ be the set of all possible 3D positions of where the target object can be. At every time step~$i$, we let the likelihood of the target object being at each point $p \in P_s$ be in the range $[0, 1]$. The target object is definitely not at point~$p$ if it has a likelihood of 0. We define $\Phi_{s, i} \subseteq P_s \times \mathbb{R}_{[0, 1]}$ as the mapping of each possible position to its likelihood in scene~$s$ at time step~$i$. Then, the lower bound estimate of the agent's uncertainty of the goal location is:
$\lambda_{s, i} = \sum_{p \in P_s} \Phi_{s, i, p} - \max_{p \in P_s}\Phi_{s, i, p}$.
We choose to subtract the maximum value of likelihood in the mapping $\Phi_{s, i}$ so as to ensure that when there is only one position with positive likelihood value, the overall uncertainty $\lambda_{s, i}$ is 0.
Beginning at time step $i=0$, the target object has equal likelihood of being at any point: 
$\forall p \in P_s: \Phi_{s, 0, p} = 1$.
It is noted that different scenes in AI2-THOR have different areas, and hence different starting overall uncertainty $\lambda_{s,0}$. We believe that this property reflects the amount of uncertainty in real life, whereby an agent would have a bigger search space if the scene is larger and vice versa.

\begin{figure*}[ht]
\includegraphics[width=1.0\linewidth]{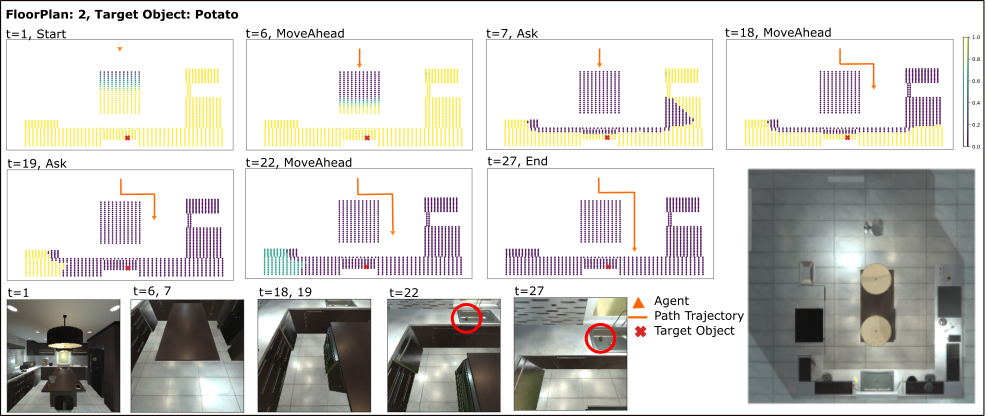}
\caption{A qualitative result for how the lower bound estimate of the agent's uncertainty of the target object's location changes: (top) rendering of the uncertainty mapping; (bottom) egocentric RGB observations at the corresponding time steps with the observable target object circled in red; (bottom-right) allocentric view of the scene.}
\label{fig:uncert}
\vspace{-10pt}
\end{figure*}

\subsubsection{Uncertainty Change from Navigation}
At each time step $i$, if the action taken is a navigation action that is not a termination (i.e. \textit{rotate left/right}, \textit{move forward} or \textit{look up/down}), then the agent can use the RGB and depth observations to decrease its uncertainty of where the target location is.
We formulate the decrease of uncertainty from this observation in two cases: when the target object is in view, and when it is not.

When the target object is not in the agent's view, the change in likelihood mapping is given by:
\begin{equation}
\label{eq:uc-nav-unseen}
    \forall p \in P_s: \Phi_{s, i, p} = \Phi_{s, i-1, p} - \psi(p) \times \textrm{in}(obs_{view}, p)
\end{equation}
where
$\textrm{in}(obs_{view}, p) = 1$ if point $p$ is in the agent's view else $0$,
and $\psi$ is a piece-wise linear decay function:
\begin{equation}
\psi(p) =
    \begin{cases}
    1 & \text{if ${dist_p} \leq \alpha$,} \\
    1 - \frac{dist_p - \alpha}{\beta - \alpha} & \text{if $dist_p \leq \beta$,} \\
    0 & \text{otherwise.}
    \end{cases}
\end{equation}
where $dist_p$ is the distance from the agent to the point $p$, and $\alpha$ and $\beta$ are pre-determined parameters. 
In our analysis, we define $\alpha=1.0$ and $\beta=2.0$, which are respectively $1\times$ and $2\times$ of the proximity distance which the agent has to be of the target object to succeed.
The decay function mimics the effect that object recognition is often better when the agent is closer to the object. For example, while an agent may mistake a tomato for an apple when it is far away, it can distinguish better when the visual details are clearer. In Equation \ref{eq:uc-nav-unseen}, the likelihood decreases for every point seen by the agent. This coincides with the agent's potential knowledge gain that the target object is not in the seen region.

Suppose that the target object is at position $\hat{p}$, then when the point $\hat{p}$ is in the agent's RGB-D view, the change in likelihood mapping is given by:
\begin{equation}
\label{eq:uc-nav-seen}
\begin{split}
\forall p \in P_s: & \Phi_{s, i, p} = \Phi_{s, i-1, p} \\
& +\;\textrm{is\_target}(p) \times \psi(p) \times \textrm{in}(obs_{view}, p) \\
& +\;\textrm{is\_target}(p) \times \psi(\hat{p})
\end{split}
\end{equation}
\begin{flalign}
&\text{where} \quad \textrm{is\_target}(p) =
    \begin{cases}
    1 & \text{if point $p=\hat{p}$,} \\
    0 & \text{otherwise.}
    \end{cases}&&
\end{flalign}
In Equation \ref{eq:uc-nav-seen}, the likelihood decreases for every point seen by the agent that does not have the target object. This reflects the agent's potential knowledge gain that the target object is not at the seen region. Then, the likelihood decreases for every point that does not have the target object, including the points not covered by the agent's view. The target object is in the agent's view, reflecting the potential knowledge gain that the target object is less likely to be at other positions.

\subsubsection{Uncertainty Change from Feedback}
At each time step $i$, if an \textit{ask} action is taken, then the additional observation that the agent can use to decrease its uncertainty of the target object's location is the object-in-view feedback. The decrease of uncertainty from this observation is formulated in two cases: when the target object is in view, and when it is not.

When the target object is not in the agent's view, the change in likelihood mapping is given by:
\begin{equation}
\label{eq:uc-ask-unseen}
    \forall p \in P_s: \Phi_{s, i, p} = \Phi_{s, i-1, p} - \textrm{in}(obs_{view}, p)
\end{equation}
As the object-in-view feedback is a ground truth observation, there is no ambiguity in it. Hence, Equation \ref{eq:uc-ask-unseen} reflects the potential knowledge gain that the current view does not contain the target object.
On the other hand, if the target object is at position $\hat{p}$ and $\hat{p}$ is in the agent's view, the change in likelihood mapping is given by:
\begin{equation}
\label{eq:uc-ask-seen}
\begin{split}
\forall p \in P_s: \Phi_{s, i, p} & = \Phi_{s, i-1, p} \\
& + \textrm{in}(obs_{view}, p) \times \textrm{is\_target}(p) \\
& + \textrm{in}(obs_{view}, p) \times \psi(\hat{p})
\end{split}
\end{equation}
The first part of Equation \ref{eq:uc-ask-seen} reflects the potential knowledge gain that positions in the current view with non-positive object-in-view feedback do not contain the target object. The second part reflects the potential knowledge gain that the target object is less likely to be at other positions.


\section{Experimental Results}

\subsection{Evaluation Metrics}
\label{sec:eval-metrics}

We evaluate our agent on these standard navigation metrics \cite{batra2020objectnav}: (1) \textbf{Success rate (SR)}, the ratio of successful episodes over completed episodes $N$: $SR = \frac{1}{N} \sum_{i=1}^{N}S_i$. (2) \textbf{Success weighted by path length (SPL)} \cite{anderson2018vision}, a measurement of the efficacy of navigation, given by: $SPL = \frac{1}{N}\sum_{i=1}^{N} S_i \times (\frac{l_i}{max(p_i, l_i)})$ where $l_i$ is the shortest path distance from the agent's starting position to the goal in episode $i$, and $p_i$ is the actual path length taken by the agent. Additionally, we propose these new evaluation metrics to justify the effectiveness of the human-agent interaction:
\begin{itemize}
    \item The percentage of actions taken that are \textit{ask} actions - how frequently the agent is asking for feedback.
    \item The average decrease in uncertainty of the target object's location from \textit{ask} actions - Using the definition in Section \ref{sec:quantifying-uncert}, we can quantify how much these \textit{ask} actions are helping the agent in the ObjectNav task.
\end{itemize}

In the real world, there is no obvious property used to quantify when is a good time to ask. However, in this controlled simulated environment, there are instances where we can point out that is not the most informative time to ask. Hence, in our analysis, we give statistics on the following types of \textit{ask} actions taken: (1) Consecutive \textit{ask} actions: The agent does not gain any observations or feedback different from previous time steps. (2) Vapid \textit{ask} actions: Ask actions taken when the agent should already have a good idea of where the goal is (i.e. the lower bound estimate of uncertainty is $<$ 10\% of its starting uncertainty). (3) Statistically insignificant \textit{ask} actions: Ask actions which decrease the uncertainty minimally (i.e. the change in lower bound estimate of uncertainty caused by the \textit{ask} action is less than a threshold $\gamma$). We choose $\gamma=2.0$, about 10\% of the average decrease in uncertainty by all actions in the baseline. These statistics are shown in Table \ref{tab:results-ask}.

\begin{table*}[t]
\centering
\resizebox{\textwidth}{!}{%
\tabulinesep=0.3mm
\begin{tabu}{clccccccccc}
\hline
 & \multicolumn{1}{c}{} & \multicolumn{4}{c}{Success Rate (\%)} & \multicolumn{1}{l}{} & \multicolumn{4}{c}{Success weighted by Path Length (\%)} \\ \cline{3-6} \cline{8-11} 
\begin{tabular}[c]{@{}c@{}}Teacher Presence\\during testing\end{tabular} & \begin{tabular}[c]{@{}c@{}}Training\\Methods\end{tabular} & \begin{tabular}[c]{@{}c@{}}Both\\Seen\end{tabular}& \begin{tabular}[c]{@{}c@{}}Both\\Unseen\end{tabular} & \begin{tabular}[c]{@{}c@{}}Unseen\\ Scenes\end{tabular} & \begin{tabular}[c]{@{}c@{}}Unseen\\Objects\end{tabular} & \multicolumn{1}{l}{} & \begin{tabular}[c]{@{}c@{}}Both\\Seen\end{tabular}& \begin{tabular}[c]{@{}c@{}}Both\\Unseen\end{tabular} & \begin{tabular}[c]{@{}c@{}}Unseen\\ Scenes\end{tabular} & \begin{tabular}[c]{@{}c@{}}Unseen\\Objects\end{tabular} \\ \hline
False & Baseline & 35.2 & 11.9 & 18.6 & 13.0 &  & 24.1 & 7.7 & 9.1 & 6.9 \\
False & Feedback & 4.8 & 0.0 & 1.3 & 0.3 &  & 1.1 & 0.0 & 0.4 & 0.1 \\
False & Semi-25 & 40.6 & 7.9 & 23.1 & 13.3 &  & 26.8 & 4.5 & 11.6 & 8.5 \\
False & Semi-75 & 33.9 & 6.7 & 18.1 & 18.5 &  & 27.4 & 6.7 & 11.4 & 16.3 \\ \hline
True & Feedback & 70.9 & \textbf{26.3} & \textbf{71.4} & \textbf{37.0} &  & 46.4 & \textbf{17.2} & \textbf{39.7} & \textbf{24.2} \\
True & Semi-25 & 51.6 & 6.6 & 35.7 & 15.1 &  & 34.1 & 4.1 & 20.2 & 8.3 \\
True & Semi-75 & \textbf{72.1} & 8.6 & 50.1 & 24.8 &  & \textbf{50.9} & 4.6  & 34.0 & 18.9 \\ \hline
\end{tabu}%
}
\caption{Success Rate (SR) and Success weighted by Path Length (SPL) evaluation results for different seen and unseen cases for objects and scenes across methods. There are a total of 4 training methods. 1) \textbf{Baseline}: the agent only has the visual (i.e. RGB-D) and object category information. 2) \textbf{Semi-25}: an agent trained with 25\% present teacher curriculum. 3) \textbf{Semi-75}: an agent trained with 75\% present teacher curriculum. 4) \textbf{Feedback}: the agent receives object-in-view feedback through object segmentation upon asking for help.}
\label{tab:results-basic}
\vspace{-5pt}
\end{table*}

\begin{table*}[t]
\centering
\resizebox{\textwidth}{!}{%
\tabulinesep=0.3mm
\begin{tabu}{clccccccccc}
\hline
 & \multicolumn{1}{c}{} & \multicolumn{4}{c}{Average change in uncertainty by Nav actions} & \multicolumn{1}{l}{} & \multicolumn{4}{c}{Average change in uncertainty by Ask actions} \\ \cline{3-6} \cline{8-11} 
\begin{tabular}[c]{@{}c@{}}Teacher Presence\\during testing\end{tabular} & \begin{tabular}[c]{@{}c@{}}Training\\Methods\end{tabular} &  \begin{tabular}[c]{@{}c@{}}Both\\Seen\end{tabular}& \begin{tabular}[c]{@{}c@{}}Both\\Unseen\end{tabular} & \begin{tabular}[c]{@{}c@{}}Unseen\\ Scenes\end{tabular} & \begin{tabular}[c]{@{}c@{}}Unseen\\Objects\end{tabular} & \multicolumn{1}{l}{} & \begin{tabular}[c]{@{}c@{}}Both\\Seen\end{tabular}& \begin{tabular}[c]{@{}c@{}}Both\\Unseen\end{tabular} & \begin{tabular}[c]{@{}c@{}}Unseen\\ Scenes\end{tabular} & \begin{tabular}[c]{@{}c@{}}Unseen\\Objects\end{tabular} \\ \hline
False & Baseline & 13.8 & 2.43 & 6.9 & 6.8 &  & - & - & - & - \\
False & Semi-25 & 15.8 & 8.8 & 8.1 & 26.1 &  & - & - & - & - \\
False & Semi-75 & 19.5 & 17.8 & 13.8 & 17.2 &  & - & - & - & - \\ \hline
True & Feedback & 15.6 & 14.2 & 6.7 & 17.3 &  & 22.2 & 26.4 & 7.6 & 21.9 \\
True & Semi-25 & 20.1 & 6.9 & 6.4 & 29.4 &  & 23.6 & 30.8 & 15.6 & 10.5 \\
True & Semi-75 & 19.4 & 7.0 & 11.6 & 11.0 &  & 28.3 & 19.9 & 21.2 & 13.6 \\ \hline
\end{tabu}%
}
\caption{Average change in agent's overall uncertainty by navigation and ask actions for different seen and unseen cases for objects and scenes across methods.}
\label{tab:results-uncert}
\vspace{-5pt}
\end{table*}

\begin{table*}[t]
\centering
\tabulinesep=0.3mm
\resizebox{\textwidth}{!}{%
\begin{tabu}{lccccccccc}
\hline
\multicolumn{1}{c}{} & \multicolumn{4}{c}{\% of Ask actions in all actions} & \multicolumn{1}{l}{} & \multicolumn{4}{c}{\% of Consecutive Ask actions in all ask actions} \\ \cline{2-5} \cline{7-10} 
\begin{tabular}[c]{@{}c@{}}Training\\Methods\end{tabular} & \begin{tabular}[c]{@{}c@{}}Both\\Seen\end{tabular}& \begin{tabular}[c]{@{}c@{}}Both\\Unseen\end{tabular} & \begin{tabular}[c]{@{}c@{}}Unseen\\ Scenes\end{tabular} & \begin{tabular}[c]{@{}c@{}}Unseen\\Objects\end{tabular} & \multicolumn{1}{l}{} & \begin{tabular}[c]{@{}c@{}}Both\\Seen\end{tabular}& \begin{tabular}[c]{@{}c@{}}Both\\Unseen\end{tabular} & \begin{tabular}[c]{@{}c@{}}Unseen\\ Scenes\end{tabular} & \begin{tabular}[c]{@{}c@{}}Unseen\\Objects\end{tabular} \\ \hline
Feedback & \textbf{34.9} & \textbf{24.6} & \textbf{42.1} & \textbf{37.4} &  & \textbf{0.3} & \textbf{0.4} & \textbf{0.2} &\textbf{ 0.4} \\
Semi-25 & 15.6 & 13.5 & 17.1 & 12.2 &  & 14.3 & 11.8 & 16.1 & 5.8 \\
Semi-75 & 23.1 & 16.2 & 19.3 & 25.8 &  & 5.4 & 5.1 & 5.5 & 5.6 \\ \hline
\multicolumn{1}{c}{} & \multicolumn{4}{c}{\% of Vapid Ask actions in all ask actions} & \multicolumn{1}{l}{} & \multicolumn{4}{c}{\% of Statistically Insignificant Ask actions in all ask actions} \\ \cline{2-5} \cline{7-10} 
Feedback & \textbf{57.8} & 70.8 & \textbf{70.4} & \textbf{63.3} &  & 80.3 & \textbf{81.2} & \textbf{94.2} & \textbf{73.8} \\
Semi-25 & 86.0 & 88.2 & 78.3 & 86.2 &  & 93.6 & 86.3 & 96.0 & 80.0 \\
Semi-75 & 75.4 & \textbf{70.1} & 73.4 & 64.6 &  & \textbf{78.9} & 83.4 & 94.3 & 87.9 \\ \hline
\end{tabu}%
}
\caption{Statistics on the types of ask actions taken for different seen and unseen cases across methods where feedback is available. Teacher is present in all episodes used for evaluation here.}
\label{tab:results-ask}
\vspace{-5pt}
\end{table*}

\begin{table*}[t]
\centering
\renewcommand{\arraystretch}{0.8}
\resizebox{\textwidth}{!}{%
\begin{tabular}{@{}ccccccccccc@{}}
\hline
 &  & \multicolumn{4}{c}{Success Rate (\%)} &  & \multicolumn{4}{c}{Success weighted by Path Length (\%)} \\ \cmidrule(lr){3-6} \cmidrule(l){8-11} 
\begin{tabular}[c]{@{}c@{}}Teacher Presence\\ during testing\end{tabular} & \begin{tabular}[c]{@{}c@{}}Training\\ Methods\end{tabular} & \begin{tabular}[c]{@{}c@{}}Both\\ Seen\end{tabular} & \begin{tabular}[c]{@{}c@{}}Both\\ Unseen\end{tabular} & \begin{tabular}[c]{@{}c@{}}Unseen\\ Scenes\end{tabular} & Unseen &  & \begin{tabular}[c]{@{}c@{}}Both\\ Seen\end{tabular} & \begin{tabular}[c]{@{}c@{}}Both\\ Unseen\end{tabular} & \begin{tabular}[c]{@{}c@{}}Unseen\\ Scenes\end{tabular} & \begin{tabular}[c]{@{}c@{}}Unseen\\ Objects\end{tabular} \\ \midrule
True & Binary Feedback & 32.8 & 22.4 & 23.7 & 18.4 &  & 22.3 & 6.1 & 9.9 & 6.3 \\
True & \begin{tabular}[c]{@{}c@{}}Feedback\\ (eval with noise)\end{tabular} & 58.8 & 18.2 & 59.0 & 20.0 &  & 36.1 & 10.5 & 33.0 & 12.2 \\
True & Language Feedback & \textbf{74.3} & \textbf{42.6} & 63.2 & \textbf{51.8} &  & 43.5 & \textbf{23.0} & 28.9 & \textbf{29.8} \\
True & Feedback & 70.9 & 26.3 & \textbf{71.4} & 37.0 &  & \textbf{46.4} & 17.2 & \textbf{39.7} & 24.2 \\ \hline
\end{tabular}
}
\caption{Success Rate (SR) and Success weighted by Path Length (SPL) evaluation results for different seen and unseen cases for objects and scenes across varied feedback signals.}
\label{tab:results-variations}
\vspace{-5pt}
\end{table*}

\subsection{Feedback Variations}

In this paper, we introduce some feedback variations: (1) \textbf{Binary Feedback}: Binary feedback flattens the object-in-view feedback into a binary signal. Upon each ask action, the agent will receive a positive scalar signal of 1 if the target object is in its view, and a scalar signal of 0 if it is not. The agent will receive a signal of -1 if the ask action is not taken. (2) \textbf{Added noise}: Agent is trained with ground truth object-in-view feedback and evaluated with noisy image semantic segmentation for a more realistic setup in the real world. The noisy image segmentation is generated from the ground truth by two perturbations. The first perturbation is to scale the ground truth segmentation vertically or horizontally within a range of 0.6 to 1.0, whereby the shape remains unchanged if scaled by 1.0. The second perturbation is to adjust the segmentation boundary by a random amount between -5 to +5 pixels. (3) \textbf{Language Feedback}: Object-in-view feedback expressed in natural language instead of image semantic segmentation, based on target object color, location and distance. The natural language inputs are generated as shown in Table \ref{tab:nlp-rules}. We use RoBERTa \cite{roberta} to encode the natural language inputs before feeding them into the policy.

\begin{table*}[t]
\centering
\tabulinesep=0.3mm
\resizebox{\linewidth}{!}{%
\begin{tabu}{|l|l|}
\hline
\textbf{Cases} & \textbf{Natural Language Feedback} \\ \hline
No ask action taken & The target object is \textless{}\textit{target object name}\textgreater{}. \\ \hline
Ask action, target object is not in view & The \textless{}\textit{target object name}\textgreater is absent from the frame. \\ \hline
Ask action, target object is in view & \begin{tabular}[c]{@{}l@{}}The \textless{}\textit{target object color}\textgreater \textless{}\textit{target object name}\textgreater is \textless{}\textit{close/ far}\textgreater{}, at the \textless{}\textit{position}\textgreater of the frame.\\\textit{position} :- \textit{top-left, top, top-right, left, middle, right, bottom-left, bottom, bottom-right}\end{tabular} \\ \hline
\end{tabu}%
}
\caption{Rules used to generate object-in-view feedback in natural language.}
\label{tab:nlp-rules}
\vspace{-10pt}
\end{table*}

\subsection{Analysis}


We use two methods to evaluate the impact of \textit{ask} actions in the agent's performance:
(1) \textbf{Baseline}: the agent only has the visual (i.e. RGB-D) and object category information.
(2) \textbf{Object-in-view feedback}: in addition to (1), the agent receives object-in-view feedback upon asking for help.
From Table \ref{tab:results-basic}, we see that object-in-view feedback method outperforms the baseline in all cases by an average of 31.7\% and 19.9\% for SR and SPL respectively. This shows that the agent understands how to use the feedback, i.e. learning the semantic meaning of the given feedback. As expected, the agent's performance decreases as more parts of the test scenario (i.e. target object and scene) are unseen. We can also see that performance deterioration is generally more significant for unseen objects than for unseen scenes and the most significant for test scenarios with simultaneously unseen objects and scenes.

Table \ref{tab:results-basic} also shows the results from our semi-present teacher training curriculum described in Section \ref{sec:semi-present-teacher}. We label Semi-$\eta$ as the method used to train an agent with $\eta$\% present teacher curriculum. We discover that the agent trained with object-in-view feedback and a 100\% present teacher fails in 95\% of all test episodes when the teacher is absent, performing significantly worse than the baseline which has a SR of at least 11.9\%. This shows that the agent is over-reliant on the teacher's feedback in a 100\% present teacher training curriculum and cannot generalize well when the teacher is absent. However, both agents trained from the 25\% and 75\% semi-present teacher curricula achieve comparable performance to that of the baseline's when the teacher is absent, with a maximum SR decrease of 5.2\%. This shows that a simple curriculum of mixing in episodes with no teacher feedback helps the agent learn to react more robustly to the absence of a teacher. The benefits of feedback can be seen as Semi-25 and Semi-75 agents achieve a maximum increase of 38.2\% in SR and 23.5\% in SPL with the presence of a teacher. 
We also note that Semi-25 and Semi-75 agents' performance with a present teacher is not as high as that when trained with an always present teacher with a minimum SR decrease of 7.7\%.

Using the uncertainty metric defined in Section \ref{sec:quantifying-uncert}, we present the average decrease in uncertainty caused by navigation or \textit{ask} actions in Table \ref{tab:results-uncert}. The average uncertainty change by \textit{ask} actions is higher than that by navigation actions for the agent trained with an always present teacher and Semi-75. This shows that the object-in-view feedback from \textit{ask} actions is statistically useful in helping the agent complete the task more efficiently.
Fig. \ref{fig:uncert} shows a qualitative example of how uncertainty changes in an episode from navigation and \textit{ask} actions.

The agent trained with an always present teacher uses the highest percentage of \textit{ask} actions.
Table \ref{tab:results-ask} shows the percentages of the different types of \textit{ask} actions taken across different training curricula.
The percentage of consecutive \textit{ask} actions taken by the agent trained with an always present teacher is close to zero, suggesting that the agent is successful in learning that there is no additional gain in information when asking consecutively.
Comparing across methods, the agent trained with an always present teacher has the lowest percentage of consecutive \textit{ask} actions, vapid \textit{ask} actions, and statistically insignificant \textit{ask} actions, followed by Semi-75 and lastly Semi-25. This suggests that while the agent trained with an always present teacher uses a higher percentage of \textit{ask} actions, it has learned to utilize them in a statistically informative manner.
While the percentages of vapid asks and statistically insignificant asks (defined in Section \ref{sec:eval-metrics}) seem high, this is partly attributed to the way the uncertainty metric is defined. This metric is a lower bound estimate of the agent's uncertainty of the goal location. It assumes a perfect memory where uncertainty only decreases with more observations, which may not be the case for an RL agent.

Table \ref{tab:results-variations} shows the results of our feedback variations. Binary feedback is a less costly form of feedback (i.e. more convenient for a user), but it has worse performance in most experimental setups with an average decrease of 46.4\% in SR and 66.4\% in SPL from object-in-view feedback. The agent that is trained with image semantic segmentation feedback has a decrease in performance for all setups when it is evaluated with noisy segmentation feedback, with the SR decreasing by an average of 12.4\% and the SPL by an average of 8.9\%. Lastly, the language feedback has the closest performance to the image semantic segmentation feedback, with an average difference of 10.7\% in SR and 6.3\% in SPL, and is also the only feedback variation to outperform the image semantic segmentation feedback in multiple setups. These results suggest that future work can focus on reducing performance degradation with noisy feedback and using natural language as a feedback modality. As future work, we plan to enable easy integration of multiple feedback modalities into the framework by implementing a multi-modal feedback module.

\section{Conclusion}
In this work, we demonstrate the importance of quality feedback signals toward robot learning for the task of object-goal navigation. We propose a learning framework that enables the embodied agent to optimize its navigation policy for ObjectNav tasks by learning to request for feedback. Additionally, we propose a semi-present teacher training curriculum to increase the agent's robustness when feedback is not always present. Finally, we develop a formulation of the lower bound estimate of the uncertainty that quantitatively measures the timing and robustness of the learning agent's feedback-seeking behavior, and show that our framework encourages effective agent behaviors for ObjectNav.

\textbf{Limitations and future work.}
(1) We use perfect feedback in our current training, which is unrealistic in a real-world setting. A future direction could be to make the interactive learning framework more robust to noisy or imperfect feedback.
(2) As a proof of concept, we briefly investigated using natural language as a feedback modal. Human interaction comes in many modals. For similar semantic meanings in a feedback, future work directions can include the study of modular and exchangeable feedback encoders whereby the algorithm does not need to be retrained for a different feedback modal.

\bibliography{main}

\end{document}